# GBSS: A GLOBAL BUILDING SEMANTIC SEGMENTATION DATASET FOR LARGE-SCALE REMOTE SENSING BULDING EXTRACTION

*Yuping Hu[1], Xin Huang[1], Jiayi Li[1,2,\*], Zhen Zhang[1]*

[1]School of Remote Sensing and Information Engineering, Wuhan University, PR China
[2]Hubei Luojia Laboratory, Wuhan University, PR China
[\*]Corresponding author: zjjerica@whu.edu.cn (J. Li)

**ABSTRACT**

Semantic segmentation techniques for extracting building footprints from high-resolution remote sensing images have been widely used in many fields such as urban planning. However, large-scale building extraction demands higher diversity in training samples. In this paper, we construct a Global Building Semantic Segmentation (GBSS) dataset (The dataset will be released), which comprises 116.9k pairs of samples (about 742k buildings) from six continents. There are significant variations of building samples in terms of size and style, so the dataset can be a more challenging benchmark for evaluating the generalization and robustness of building semantic segmentation models. We validated through quantitative and qualitative comparisons between different datasets, and further confirmed the potential application in the field of transfer learning by conducting experiments on subsets.

## 1. INTRODUCTION

The geographical information of buildings, such as location and area, plays an irreplaceable role in various applications, including population estimation 0, urban planning [2], disaster assessment [3], land use analysis [4] and map updating [5]. The spatial resolution of remote sensing images has seen a remarkable increase since imaging technology developed, ushering in the era of Very High-Resolution (VHR) imagery, which enables rapid and accurate large-scale (e.g., global) building extraction. However, large-scale building extraction still faces great challenges in data.

Although many high-precision and high-efficiency methods have been proposed, however, the scarcity of large-scale datasets has hindered further advancements in the methodology. Manual annotation of pixel-level building labels demands a substantial amount of human effort, and this high cost leads to the scarcity of large-scale datasets. The existing datasets still lack sample diversity, making it challenging to measure the generalization performance of methods. The commonly used building semantic segmentation datasets often cover only single or a few cities. For example, the WHU building dataset only has samples from Christchurch [6], resulting in building and background similarities among different samples.

In this paper, we constructed a Global Building Semantic Segmentation (GBSS) dataset in a semi-automated manner. To acquire building segmentation samples at a global scale, we overlay open-source vector data OpenStreetMap (OSM) and Google Maps with Global Impervious Surface Analysis (GISA) product [7] as prior knowledge. Then we developed a human-machine interactive building sample collection software to select high-quality samples, which make up the final dataset. The advantages of the dataset lie in: a) large sample size for adequate training, b) rich sample diversity for improving generalization performance, and c) wide geographical coverage for transfer learning applications.

## 2. HIGH-RESOLUTION REMOTE SENSING BUILDING SEGMENTATION DATASETS

Table. 1 lists the characteristics of mainstream high-resolution remote sensing building segmentation datasets. The WHU aerial dataset [6] and Massachusetts dataset [8] only include samples from a single city, leading to highly homogeneous building samples. Though Inria dataset [9] and SpaceNet 1/2 dataset [10] cover more representative cities, the total number of samples is still less than 20,000 if cropped to a size of 512×512.

The ISPRS-Vaihingen/Potsdam dataset [11] primarily covers the corresponding two cities and their surrounding areas. Due to computational limitations, images are generally cropped before being fed into the semantic segmentation network. However, these two datasets have very high resolutions (<0.1m), and the cropped images may not encompass entire buildings, which is disadvantageous for learning the shape of buildings and the structural relationships within building clusters.

In contrast, the GBSS dataset features a resolution of 0.25m, more than five times the sample size of the aforementioned datasets, and global coverage, rendering it significantly advantageous for large-scale extraction tasks.

**Table. 1.** Characteristics comparisons with open-source high-resolution remote sensing building segmentation datasets

| Dataset | Resolution | Size | Coverage | Data format | |
|---|---|---|---|---|---|
| | | | | Image | Label |
| WHU (aerial) | 0.3m | 22,000 buildings<br>8,189 samples<br>(512×512) | Christchurch | RGB | raster |
| Massachusetts | 1m | 151 samples<br>(1500×1500) | Boston | RGB | Shp,raster |
| Inria | 0.3m | 180 samples<br>(5000×5000) | Austin, Chicago, Kitsap County, Western Tyrol, Vienna | RGB | Raster |
| SpaceNet 1/2 | 0.5m/0.3m | 6940 samples<br>(438 × 406)<br>10,593 samples<br>(650 × 650) | Rio de Janeiro, Las Vegas, Paris, Shanghai, Khartoum, Atlanta | RGB/8-Band | GeoJSON |
| ISPRS-Vaihingen | 0.09m | 33 samples<br>(7680×13824) | Vaihingen | RG-NIR+DSM | Raster |
| ISPRS-Potsdam | 0.05m | 38 samples<br>(6000×6000) | Potsdam | RGB-NIR+DSM | Raster |
| GBSS | 0.25m | about 742k buildings<br>116.9k samples<br>(512×512) | Africa, Asia, Australia, Europe, South America, Australia | RGB | Raster |

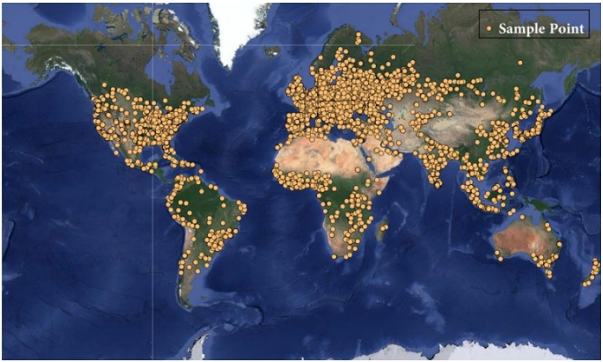

**Fig. 1.** Geographic distribution of GBSS dataset samples.

### 3. GBSS DATASET

#### 3.1. Data specification

We constructed a global satellite imagery dataset for building segmentation, named Global Building Semantic Segmentation (GBSS) dataset. The imagery and building labels are sourced from Google Maps and OpenStreetMap respectively. The dataset covers six continents, including Asia, Africa, Europe, Australia, North America, and South America, spanning a total area of about 1310 km$^2$ and containing approximately 742,000 building instances. The image of each sample has RGB three-band channels with a size of 512 × 512 and a resolution of 0.25m, and the corresponding label is a binary raster map of the same resolution. As shown in Fig. 1, the samples are mostly distributed in Europe, North America and Asia, along with the south-central Africa and developed coastal areas in Australia and South America. All 116.9k pairs of samples are divided into training set, verification set and test set in a ratio of approximately 5:1:1.

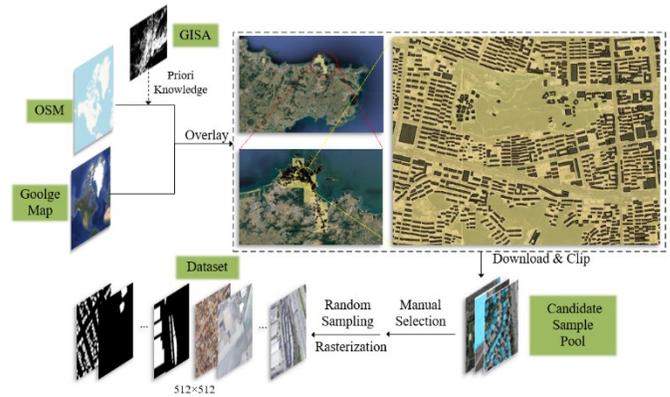

**Fig. 2.** GBSS dataset production flowchart. In the upper right sub-figure, the yellow grid represents regions with dense impervious surface area and a high number of buildings, and the black polygons represent the OSM building vectors.

#### 3.2. Production process

We overlaid building vectors from OpenStreetMap (OSM) and Google satellite imagery to create a rich and diverse collection of building samples spanning across six continents, jointly utilizing the 30-m Global Impervious Surface Analysis (GISA) product [7] as prior knowledge. The data production process is illustrated in Fig. 2.

*Step 1*: Candidate sample extraction on the Google Earth Engine platform. The high-density impervious surface often indicates the presence of numerous buildings, as building is an important subcategory of impervious surface. Therefore, the impervious surface ratio and the number of buildings can serve as two indicators of the completeness of OpenStreetMap (OSM) annotations. We calculated these two indicators within each 4km × 4km non-overlapping window relying on GISA product. Empirical thresholds were defined to identify potential sampling areas, then we non-overlappingly cropped them into 512×512 patches to create a candidate sample pool.

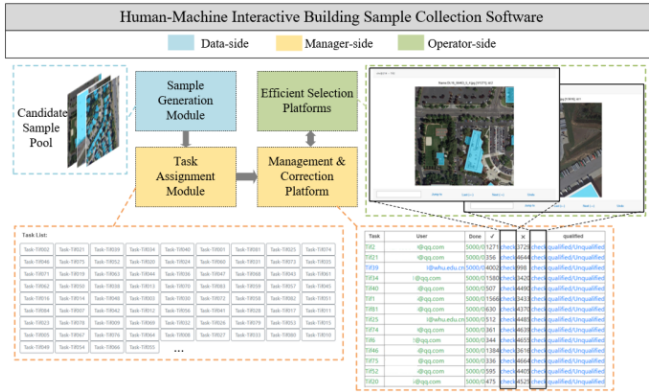

**Fig. 3.** Human-machine interactive building sample collection software.

*Step 2*: High-quality sample selection through human-machine interaction. Affected by the quality of OSM data, the candidate sample pool may still contain numerous unusable samples with label errors, such as missing, misaligned, or incorrectly shaped. To address this, we developed a human-machine interactive building sample collection software to eliminate these low-quality samples (as illustrated in Fig. 3). On the data-side, the candidate sample pool was generated through Step 1. On the manager-side, the candidate samples were divided into different tasks, which were then assigned to multiple operators for selection in parallel. Operators needed to select samples with minimal misalignments, missing, or other errors in labels. Meanwhile, the manager-side arranged for professionals to check the selection results. The bottom right screenshot in Fig. 3 displayed the schedule management interface on the manager-side. The blue and green color indicated tasks that were worked upon or completed respectively. In total, about 60 professionally trained remote sensing image interpreters worked for four months to complete all tasks. The retained high-quality samples constituted the preliminary global building dataset, with all building vectors converted into raster data paired with 0.25m resolution imagery.

### 3.3. Sample diversity across continents

#### 3.3.1. Diversity of Building Sizes

In building extraction, small buildings are prone to being missed, while large buildings tend to be over-segmented. A large percentage of both types of buildings mentioned above in the dataset will increases the omission rate of the extraction results [12]. For GBSS dataset, we analyzed the proportion of buildings of different sizes across continents, following the building sizes classification criteria proposed by [13], as shown in Fig. 4. In General, small and medium-sized buildings are predominant in all continents, with medium-sized buildings having the highest proportion and large-sized buildings having the lowest. Africa and Asia have a relatively higher proportion of small-sized buildings, North America has more medium-sized buildings, and Australia, Europe, and South America have a greater number of large-sized buildings, especially in Europe.

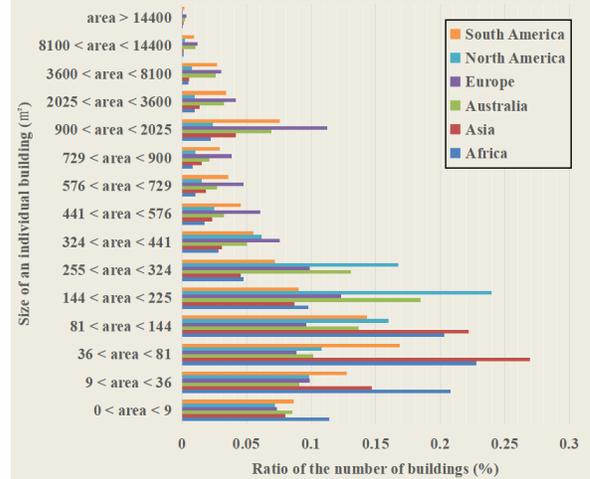

**Fig. 4.** Building size distributions for the GBSS dataset in different continents.

#### 3.3.2. Diversity of Building Styles

Apart from the diversity of building sizes, building styles such as shape and distribution characteristics also varies a lot in different geographical regions. Fig. 5 illustrates several samples from different continents. In Africa, buildings are predominantly low-rise and small to medium-sized with simple rectangular shapes, usually arranged in compact and orderly clusters. In Asia, the building shapes are more complex and numerous contiguous residential areas can be found with its high population density. In the remaining continents, Australia, Europe, and South America are characterized by a prevalence of low-rise buildings and also have a considerable number of complex-shaped medium to large-sized buildings as Asia, demanding more precise building shape extraction. In contrast, buildings in North America shows more structured distribution, with grasslands or flat terrains as the common background.

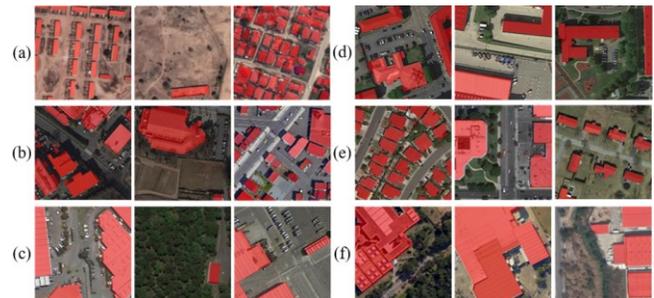

**Fig. 5.** Examples of samples from different continents. (a) Africa, (b) Asia, (c) Australia, (d) Europe, (e) North America, (f) South America.

## 4. EXPERIMENTS AND RESULTS

### 4.1. Implement details

All experiments were conducted using PyTorch on an NVIDIA GeForce RTX 2080Ti GPU with 12 GB of memory. We employed DeepLabV3+ [14], which performs best in the DeepLab series. The model is optimized by an AdamW optimizer with a learning rate of 0.00012, a weight decay of 0.01, and $\beta_1$, $\beta_2$ were set to default values, i.e., 0.9 and 0.999. Additionally, we applied a polynomial decay learning rate strategy with a power value of 0.9. The standard cross-entropy loss was utilized as the loss function. The batch size was set to 2. All models were trained from scratch until the validation accuracy no longer increases within 300k iterations or reached the maximum iteration number of 3000k. Moreover, two data augmentation techniques, random flipping and photometric distortion, were employed during training. The evaluation metrics include Intersection over Union (IoU), precision, recall and F1-score.

### 4.2. Comparison with other open building datasets

In this section, we evaluate the same segmentation network DeepLabV3+ on our GBSS dataset and two other open datasets, i.e., the WHU dataset and the Potsdam dataset. We adopted the lightweight MobileNetV2 [15] as the backbone to enhance the efficiency of large-scale building extraction.

We conducted training and testing for each dataset separately. The accuracy on the GBSS dataset is lower than the other two datasets, as shown in Table. 2. This is due to the fact that the samples of WHU dataset originate from a single city, with a relatively homogeneous building style, and mostly small and medium-sized buildings arranged densely and neatly. Thus, the network is more likely to learn the structural features of building distribution (as illustrated in Fig. 6). For Potsdam dataset, higher resolution images enable a clearer depiction of ground details, but the cropped samples have many incomplete buildings, which is not conducive to learning structural information in large-scale building extraction. In contrast, our GBSS dataset extends far beyond the scope of a city, covering a much wider geographical range. As a result, the samples exhibit greater diversity in building sizes and shapes, requiring higher level of model generalization capability for accurate extraction.

**Table. 2.** Comparison of semantic segmentation results on GBSS dataset, WHU dataset and Potsdam dataset using the same model.

| Dataset | IoU/% | Precision/% | Recall/% | F1-score/% |
|---|---|---|---|---|
| GBSS | 67.06 | 85.61 | 75.58 | 80.28 |
| WHU | 82.50 | 91.25 | 89.59 | 90.41 |
| Potsdam | 76.48 | 90.69 | 83.00 | 86.68 |

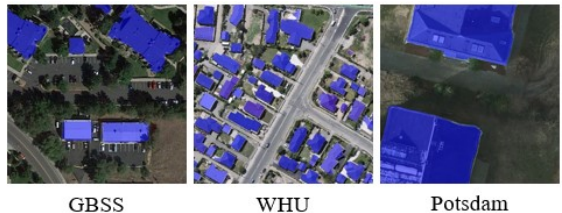

**Fig. 6.** Visualization examples of segmentation results (blue) on the three datasets.

### 4.3. Comparison between sub-datasets

We selected MobileNetV2 [15] and ResNet101 [16] as backbone to evaluate the performance of lightweight and non-lightweight models across different continental subsets within the GBSS dataset. As shown in Table. 3, relatively higher accuracy was observed in Australia, Europe, and North America subsets. This is primarily because low-rise building extractions are less affected by shadows and building facades. Additionally, the backgrounds mainly comprising grasslands and trees, distinctly different from the architectural features, facilitated clearer differentiation. The remaining three continents faced more formidable challenges. Africa exhibited a more cluttered background. Asia had a significant number of high-rise buildings. South America had a limited sample count. The disparities among subsets from diverse geographical regions further present possibilities for the application of GBSS dataset in transfer learning research.

**Table. 3.** IoU(%) on GBSS sub-datasets in different regions.

| Region | DeepLabV3+ (MobileNetV2) | DeepLabV3+ (ResNet101) |
|---|---|---|
| Africa | 51.36 | 57.25 |
| Asia | 57.69 | 60.80 |
| Australia | 78.32 | 80.47 |
| Europe | 74.68 | 76.45 |
| North America | 75.17 | 80.09 |
| South America | 59.41 | 61.59 |
| Global | 67.06 | 70.12 |

## 5. CONCLUSION

In this paper, we introduce a Global Building Semantic Segmentation (GBSS) dataset, which comprises 116.9k pairs of samples (about 742k buildings) from six continents. We discuss the characteristics of the GBSS dataset and compare it with other open-source building datasets, to prove that it can serve as a strong benchmark for large-scale (e.g., global) building extraction with such abundant sample diversity and extensive sample size. In addition, the dataset can also be used in the research of transfer learning methods. Based on this benchmark, we will continue to design a building extraction method with strong generalization performance.

## 6. ACKNOWLEDGMENTS

This work was supported in part by the Special Fund of Hubei Luojia Laboratory under Grant 220100031, and in part by the Wuhan 2022 Dawning under Project 20220108010201023.